%% file: main.tex
\documentclass[runningheads]{llncs}
\usepackage{graphicx}
\usepackage{framed,multirow}
\usepackage{amssymb}
\usepackage{url}
\usepackage[dvipsnames]{xcolor}
\usepackage{amsmath}
\usepackage{latexsym}
\usepackage{booktabs}
\usepackage{graphics}
\usepackage{graphicx}
\newcommand{\DATASET}{ArtGallery3D}
\newcommand{\DATASETS}{AG3D}
\newcommand{\POINTNAV}{PointNav$\mathcal{++}$}

\newcommand{\ie}{\textit{i}.\textit{e}.,}

\newcommand{\tit}[1]{\smallbreak\noindent\textbf{#1.}}

\newcommand{\tinytit}[1]{\noindent\textbf{#1.}}

\begin{document}
%
\title{Embodied Navigation at the Art Gallery}
%
%
\author{
Roberto Bigazzi\orcidID{0000-0002-6457-1860} \and
Federico Landi\orcidID{0000-0003-2092-1934} \and \\
Silvia Cascianelli\orcidID{0000-0001-7885-6050} \and 
Marcella Cornia\orcidID{0000-0001-9640-9385} \and
Lorenzo Baraldi\orcidID{0000-0001-5125-4957} \and 
Rita Cucchiara\orcidID{0000-0002-2239-283X}
}
%
%
\authorrunning{Bigazzi et al.}
\institute{University of Modena and Reggio Emilia, Modena, Italy\\
\email{\{name.surname\}@unimore.it}}
\maketitle              
\begin{abstract}
Embodied agents, trained to explore and navigate indoor photorealistic environments, have achieved impressive results on standard datasets and benchmarks. So far, experiments and evaluations have involved domestic and working scenes like offices, flats, and houses. In this paper, we build and release a new 3D space with unique characteristics: the one of a complete art museum. We name this environment \DATASET~(\DATASETS). Compared with existing 3D scenes, the collected space is ampler, richer in visual features, and provides very sparse occupancy information. This feature is challenging for occupancy-based agents which are usually trained in crowded domestic environments with plenty of occupancy information.
Additionally, we annotate the coordinates of the main points of interest inside the museum, such as paintings, statues, and other items. Thanks to this manual process, we deliver a new benchmark for PointGoal navigation inside this new space. Trajectories in this dataset are far more complex and lengthy than existing ground-truth paths for navigation in Gibson and Matterport3D. We carry on extensive experimental evaluation using our new space for evaluation and prove that existing methods hardly adapt to this scenario. As such, we believe that the availability of this 3D model will foster future research and help improve existing solutions.

\end{abstract}
\section{Introduction}
\label{sec:introduction}
\input{sections/01_introduction}
\vspace{-0.1cm}

\section{Related Work}
\label{sec:related}
\input{sections/02_related}
\vspace{-0.1cm}

\section{\DATASET~(\DATASETS) Dataset}
\label{sec:dataset}
\input{sections/03_dataset}
\vspace{-0.1cm}

\section{Architecture}
\label{sec:architecture}
\input{sections/04_method}
\vspace{-0.15cm}

\section{Experiments}
\label{sec:experiments}
\input{sections/05_experiments}
\vspace{-0.2cm}

\section{Conclusion}
\label{sec:conclusion}
\input{sections/06_conclusion}
\vspace{-0.2cm}

\section*{Acknowledgment}
This work has been supported by ``Fondazione di Modena'' and the ``European Training Network on PErsonalized Robotics as SErvice Oriented applications'' (PERSEO) MSCA-ITN-2020 project.
%
%
%
\bibliographystyle{splncs04}
\bibliography{bibliography}
\end{document}

%% file: sections/01_introduction.tex
In recent years, Embodied AI has benefited from the introduction of rich datasets of 3D spaces and new tasks, ranging from exploration to PointGoal or ImageGoal navigation~\cite{chang2017matterport3d,xia2018gibson}. Such availability of 3D data allows to train and deploy modular embodied agents, thanks to powerful simulation platforms~\cite{savva2019habitat}. Despite the high number of available spaces, though, the topology and nature of the different scenes have low variance. Indeed, many environments represent apartments, offices, or houses. In this paper, we take a different path and collect and introduce the 3D space of an art gallery.

Current agents for embodied exploration feature a modular approach~\cite{bigazzi2022impact,chaplot2019learning,ramakrishnan2020occupancy}. While the agents are trained for embodied exploration using deep reinforcement learning, this hierarchical paradigm allows for great adaptability on downstream tasks. Hence, models trained to explore the Gibson dataset can solve PointGoal navigation with satisfactory accuracy under the appropriate hypotheses. Furthermore, accurate and realistic simulating platforms such as Habitat~\cite{savva2019habitat} facilitate the deployment in the real world of the trained agents~\cite{bigazzi2021out,kadian2020sim2real}. While agent architectures and simulating platforms are possible sources of improvement, there is a third important direction of research that regards the availability of 3D scenes to train and test the different agents. Indeed, the nature of the different environments influences the variety of tasks that the agent can learn and perform. 

In this work, we contribute to this third direction by collecting and presenting a previously unseen type of 3D space, \ie~a museum. This new environment for embodied exploration and navigation, named~\DATASET~(\DATASETS), presents unique features when compared to flats and offices. First, the dimension of the rooms drastically increases, and the same goes for the size of the building itself. In our 3D model, some rooms are as big as $20 \times 15$ meters, while the floor hosting the art gallery spans a total of 2,000 square meters. However, dimensions are not the only difference with current available 3D spaces. As a second factor, the presented gallery is incredibly rich in visual features, offering multiple paintings, sculptures, and rare objects of historical and artistic interest. Every item represents a unique point of interest, and this is in contrast to traditional scenes where all elements have approximately the same visual relevance. Finally, the museum has sparse occupancy information. Many agents count on depth information to plan short-term displacements. However, when placed in the middle of an open empty hall, depth information is less informative. In our challenging 3D scene, the agent must learn to combine RGB and depth information and not be overconfident on immediately available knowledge on the occupancy map. All these challenges make our newly-proposed 3D space a valuable asset for current and future research.

Together with the 3D model of the museum, we present a dataset for embodied exploration and navigation. For the navigation task, we annotate the position of most of the points of interest in the museum. Examples include numerous paintings, sculptures, and other relevant objects. Finally, we present an experimental analysis including the performance of existing architectures on this novel benchmark and a discussion of potential future research directions made possible by the presence of the collected 3D space.

%% file: sections/02_related.tex
Both autonomous robotics~\cite{bigazzi2022impact,irshad2021hierarchical} and embodied AI~\cite{cascianelli2018full,chen2019learning,niroui2019deep,bigazzi2020explore,cornia2020smart,ramakrishnan2021exploration} have recently witnessed a boost of interest, which has been enabled by the release of photorealistic 3D simulated environments. In such environments, algorithms for intelligent exploration and navigation can be developed safely and more quickly than in the real-world, before being easily deployed on real robotic platforms~\cite{kadian2020sim2real,bigazzi2021out,anderson2021sim}. Among the datasets of spaces, the most commonly used are MP3D~\cite{chang2017matterport3d}, Gibson~\cite{xia2018gibson}, HM3D~\cite{ramakrishnan2021hm3d}, and Replica~\cite{straub2019replica}. These datasets mainly contain house-like and office-like environments, with some environments taken from shops, garages, churches, and restaurants. Rooms in such environments are generally cluttered, and thus, rich of landmarks and texture information that the agent can exploit while navigating. In contrast, the presented \DATASETS~environment has been collected in a museum, with larger, uncluttered spaces.

Algorithms developed in the simulated environments are typically trained with deep reinforcement learning, both for exploration and navigation tasks. The exploration task, which is often tackled to allow other downstream navigation tasks~\cite{ye2021auxiliaryON,ye2021auxiliaryPN}, consists in letting an agent equipped with visual sensors (\ie~RGB-D cameras) freely navigate the environment to gather as much information as possible, usually in the form of an occupancy map. To this end, intrinsic rewards have been proposed, which can be based on novelty, curiosity, reconstruction enabling, and coverage~\cite{ramakrishnan2021exploration,chen2019learning,ramakrishnan2020occupancy}. 
For the navigation tasks, the agent is deployed in an unknown environment (\ie~no map provided) and given some assignments in visual or textual form. These tasks include PointGoal navigation~\cite{anderson2018evaluation}, where the robot is expected to reach a coordinates-specified goal, ImageGoal navigation~\cite{zhu2017target}, where the robot must reach an observation point in the environment that matches an image-specified goal, and ObjectGoal navigation~\cite{batra2020objectnav}, where the robot is asked to get to any instance of a label-specified object in the environment. Other related tasks involve embodied question answering~\cite{das2018embodied} and vision-and-language navigation~\cite{anderson2018vision,krantz2020beyond,landi2021multimodal}, where the robot must follow a natural language instruction to reach the goal. The environment presented in this paper is used for the exploration task and for the PointNav task. In this latter case, we define a variant in which the goal is expressed in terms of both coordinates and orientation.

%% file: sections/03_dataset.tex
Existing datasets for indoor navigation comprise 3D acquisitions of different types of buildings, ranging from private houses, that cover the majority of the scenes, to offices and shops. Nevertheless, the focus of these datasets is on private spaces and there is low variance in terms of dimension and contained objects. In fact, to the best of our knowledge, among the publicly available datasets, no acquired indoor environment is composed of large rooms with a low occupied/free space ratio as in a museum. To overcome this deficiency in current literature we release a new indoor dataset for exploration and navigation captured inside a museum environment, called \DATASETS\footnote{The dataset has been collected at the Galleria Estense museum of Modena and can be found at \url{https://github.com/aimagelab/ag3d}.}.

\begin{figure}[t!]
\centering
\includegraphics[width=0.95\linewidth]{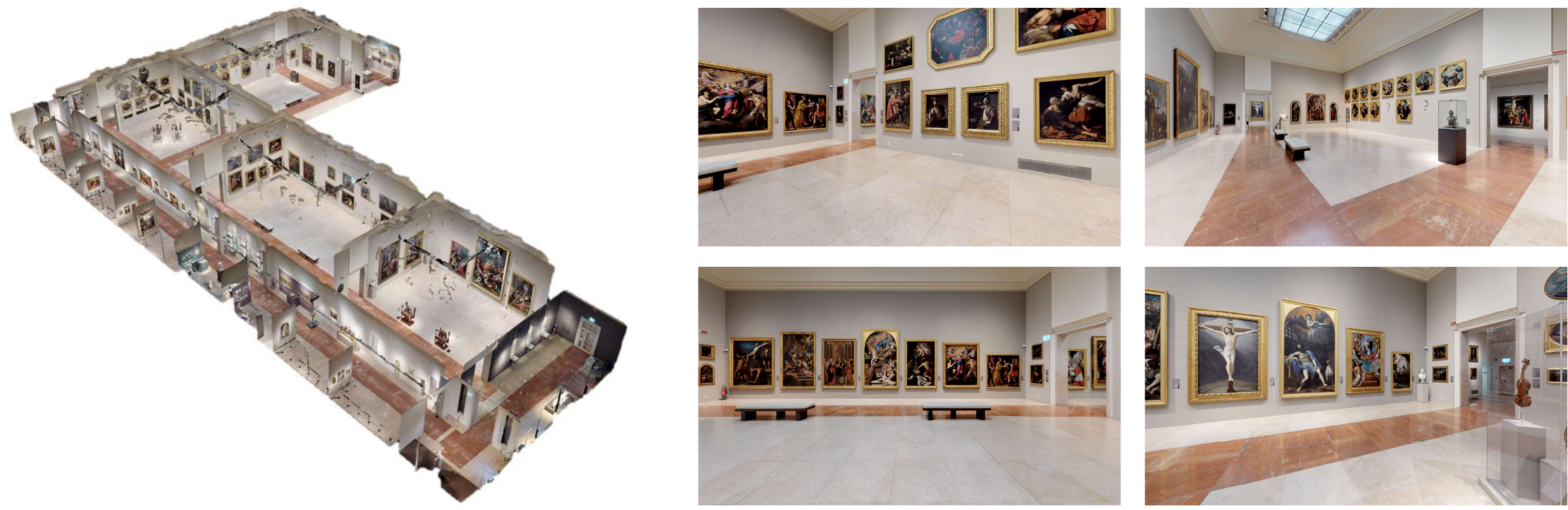}
\vspace{-0.2cm}
\caption{On the left: a view of the 3D model of the acquired environment. On the right: images captured during the acquisition of the scene.}
\label{fig:gallerie}
\vspace{-0.3cm}
\end{figure}

\tit{Acquisition}
To build the 3D model of the art gallery, we employ a Matterport camera\footnote{\url{https://matterport.com/it/cameras/pro2-3D-camera}} and related software. This technology is the same employed to collect Matterport3D and HM3D datasets of spaces~\cite{chang2017matterport3d,ramakrishnan2021hm3d} and is particularly suitable to capture indoor photorealistic environments.
We place the camera in the physical environment and capture a 360\textdegree~RGB-D image of the surrounding. Then, we repeat the same process after moving the camera approximately 1.5 meters away. Using consecutive panoramic acquisitions, the software is able to compute the 3D geometry of the space using depth information and the correspondences between the same keypoints in different acquisitions.
To capture the entire museum, we make 232 different scans. Thanks to the high number of acquisitions, we are able to reproduce fine geometric and visual details of the original space (see Fig.~\ref{fig:gallerie}). The resulting 3D model consists of more than $1430$~m\textsuperscript{2} of navigable space.

\begin{figure}[t!]
\centering
\includegraphics[width=0.8\linewidth]{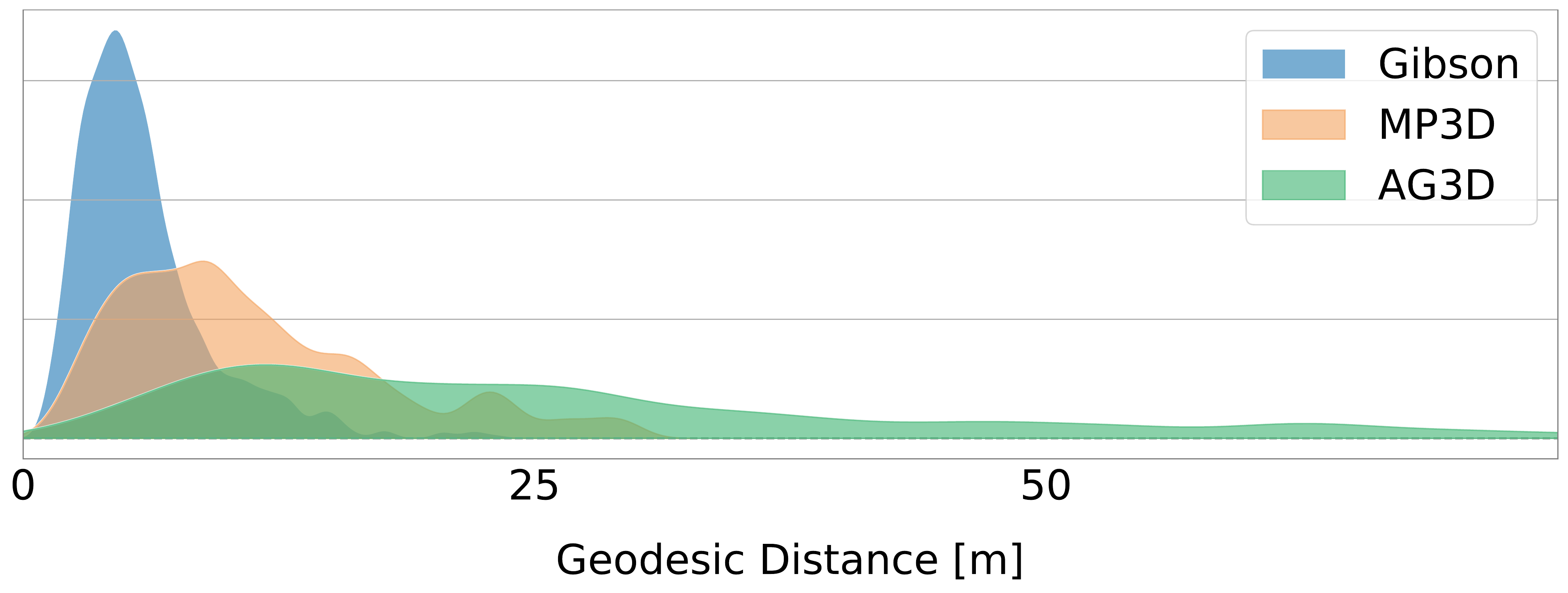}
\vspace{-0.2cm}
\caption{Comparison of the distribution of the geodesic distances from starting position to goal position of the episode for different datasets.}
\label{fig:geodesic_distances}
\vspace{-0.3cm}
\end{figure}

\tit{Dataset details}
The proposed dataset allows two different tasks: exploration and navigation. Episodes for the exploration task include starting position and orientation of the agent which are sampled uniformly over the entire navigable space. The navigation dataset, instead, extends traditional PointGoal navigation where episodes are defined with a starting pose and a goal coordinate, including an additional final orientation vector. Conceptually, we can consider this setting as the link between PointGoal navigation and ImageGoal navigation since the goal is to rotate the agent towards a precise objective/scene, specifying the goal using coordinates instead of an image. We name this new setting PointGoal$\mathcal{++}$ navigation (\POINTNAV). To create the navigation dataset we annotate 147 points of interest mostly consisting of paintings and statues. The annotated goal position is around 1 meter in front of the artwork and the goal orientation vector is directed to its center. For each point of interest, we define three episodes with different difficulties based on the geodesic distance between start and goal positions: easy ($<15$m), medium ($>15$m), and difficult ($>30$m). In particular, thanks to the dimension of the acquired environment, each difficult episode has a geodesic distance larger than the longest path of MatterPort3D and Gibson datasets. A comparison of the geodesic distance distribution of the episodes of various available PointGoal navigation datasets is presented in Fig.~\ref{fig:geodesic_distances}. The introduction of \DATASETS~enables the evaluation of agents on long navigation episodes which were previously not possible and highlights the inaccuracy of components of the architecture that accumulate error over time.
The exploration task dataset contains 500k, 100, 1000 episodes respectively for training, validation, and test, while the \POINTNAV~dataset includes 411 annotated navigation episodes.

%% file: sections/04_method.tex
\begin{figure}[t!]
\centering
\includegraphics[width=0.98\linewidth]{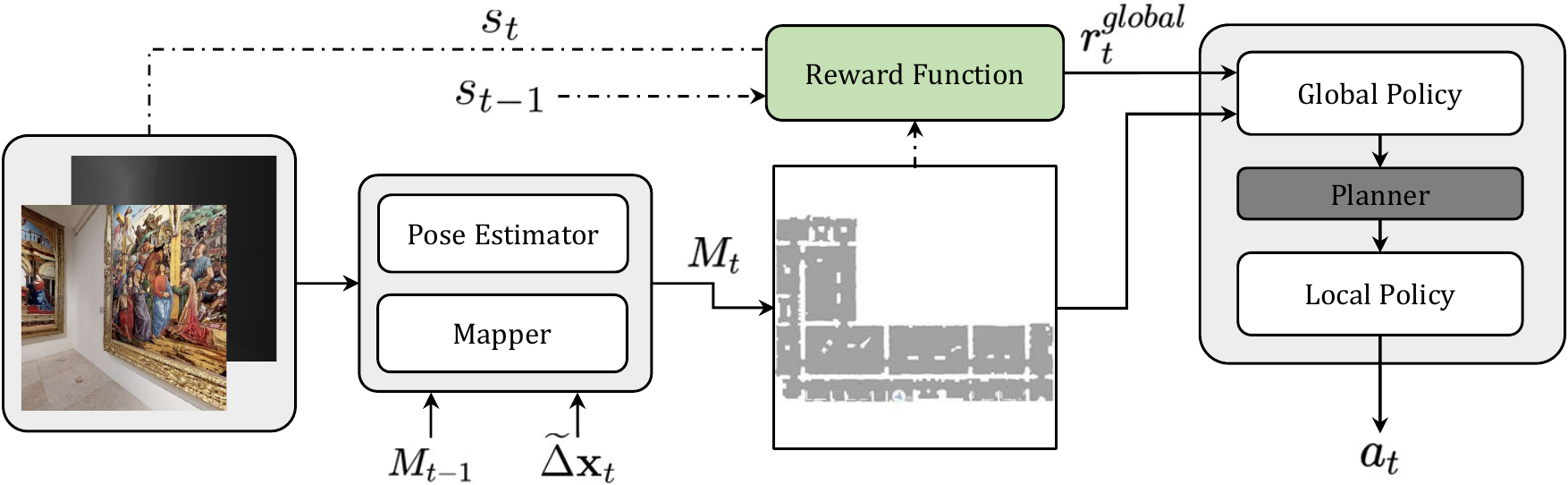}
\vspace{-0.2cm}
\caption{Overall architecture of the models employed for exploration and navigation on \DATASETS.}
\label{fig:architecture}
\vspace{-0.3cm}
\end{figure}

We provide an experimental analysis comparing recently proposed approaches on the devised environment, both for exploration and \POINTNAV~tasks. The evaluated methods are consistent with recent literature on embodied AI~\cite{chaplot2019learning,ramakrishnan2020occupancy,bigazzi2022impact} and adopt an architecture shown in Fig.~\ref{fig:architecture}, which is composed of a neural mapper, a pose estimator, and a hierarchical navigation policy.
The mapper generates a representation of the environment while the agent moves, the pose estimator is in charge of locating the agent in the environment, and the policy is responsible for the movement capabilities of the agent. The core difference between the evaluated approaches resides in the navigation policy, as described in the following. For further details, we refer the reader to the original papers.

\subsection{Mapper}
The mapper module incrementally builds an occupancy grid map of the environment in parallel with the navigation task. At each timestep, the RGB-D observations $(s^{rgb}_t, s^d_t)$ coming from the visual sensors are processed to extract a $L \times L \times 2$ agent-centric map $m_t$ where the channels indicate, respectively, the occupancy and exploration state of the currently observed region, and each pixel of the map describes the state of an area of $5 \times 5$ cm. The RGB observation is encoded using a ResNet-18 followed by a UNet, while the depth observation is encoded using another UNet. The features extracted from the two modalities are combined using CNNs at different levels of the output of the two UNet encoders and a final UNet decoder is used to process the combined features to retrieve the resulting local map $m_t$. Following the method proposed in~\cite{ramakrishnan2020occupancy}, our mapper is not limited to predicting the occupancy map of the visible space but tries to infer also occluded and not visible regions of the local map.
The global level map of the environment $M_t$ has a dimensionality of $G \times G \times 2$, where $G > L$, and is built using local maps $m_t$ step-by-step. At each timestep the pose of the agent $x_t$ is used to apply a rototranslation to the local map, then, the transformed local map is finally registered to the global map $M_t$ with a moving average.

\subsection{Pose Estimator}
In order to create a coherent representation of the environment during navigation, a precise and robust pose estimation needs to be achieved. To address problems like noise in the sensors and collisions with obstacles, we adopt a pose estimator and avoid the direct use of sensor readings. The pose estimator computes the pose of the agent $\mathbf{x}_t = (x_t, y_t, \theta_t)$ where $(x_t, y_t)$ and $\theta_t$ are its position and orientation in the internal representation of the environment. The pose estimate $\mathbf{x}_t$ is computed in an incremental way, adding the displacement $\Delta \mathbf{x}_t$ caused by the action $a_t$ to the current pose estimate $\mathbf{x}_t$. In order to retrieve a first noisy estimate of the displacement $\widetilde{\Delta}\mathbf{x}_t$, we use the difference between consecutive readings of the pose sensor $(\widetilde{\mathbf{x}}_{t-1},\widetilde{\mathbf{x}}_t)$. To account for errors in the sensor displacement, consecutive local maps $(m_{t-1},m_{t})$ coming from the mapper are used as feedback; $m_{t-1}$ is reprojected using $\widetilde{\Delta}\mathbf{x}_t$ to the same point of view of $m_t$ and the concatenation of the transformed $m_{t-1}$ and $m_{t}$ is processed using a CNN to retrieve the final robust pose displacement $\Delta \mathbf{x}_t$. At each timestep $\Delta \mathbf{x}_t$ is used to compute the pose of the robot $\mathbf{x}_{t}$:

\begin{equation}
    \mathbf{x}_t = \mathbf{x}_{t-1} + \Delta \mathbf{x}_t,
    \label{eq:estimate_displacement}
\end{equation}
where we assume $\mathbf{x}_0 = (0, 0, 0)$ without loss of generality and $\mathbf{x}_0$ corresponds to the center of the map $M_t$ with the agent facing north.

\subsection{Navigation Policy}
The navigation policy is the module that determines the movement of the agent in the environment. Its hierarchical design is required in order to allow the agent to uncouple high-level navigation concepts, such as moving across different rooms, and low-level concepts, like obstacle avoidance. The navigation policy is defined by a three-component module consisting of a global farsighted policy, a deterministic planner, and a local policy for atomic action inference.

\tit{Global Policy} The global policy is the high-level component of the navigation policy and is responsible for extracting a long-term goal on the global map $g_t$. The global policy takes as input an enriched $G \times G \times 4$ current global map $M^+_t$ retrieved stacking the two-channel global map $M_t$, the one-hot representation of the current position on the map, and the map of the already visited states. $M^+_t$ is in parallel cropped with respect to the position of the agent and max-pooled to a lower dimensionality $H \times H \times 4$. These two versions of $M^+_t$ are stacked together to obtain the final $H \times H \times 8$ input of the global policy. A CNN is used to sample a point of a $H \times H$ grid that is converted to a goal position on the global map $g_t$.
The global policy is trained with reinforcement learning using PPO~\cite{schulman2017proximal} to maximize different rewards used in literature. 

In the experiments, we employ and compare different reward methods, namely Coverage, Anticipation, and Curiosity. The Coverage reward~\cite{chaplot2019learning,ramakrishnan2021exploration} maximizes the information gathered at each time-step, expressed in terms of the area seen. The Anticipation reward~\cite{ramakrishnan2020occupancy} is defined by comparing the predicted local occupancy map with the ground-truth considering also occluded areas. The Curiosity reward~\cite{pathak2017curiosity} encourages the agent towards areas that maximize the prediction error of a model trained to predict future states, thus improving the learning of the dynamics of the environment.

\tit{Planner} Given the global goal on the map, the planner has the task of computing a short-term goal on the map that the agent should reach. We employ an A* algorithm on the global map $M_t$ to plan a path from the current position of the agent to the global goal and a local goal $l_t$ is computed on the obtained trajectory within a distance $D$ from the agent.

\tit{Local Policy} The local policy is the module that allows the movement of the agent in the environment and its objective is to reach the local goal $l_t$ determined by the planner. The input of the local policy, formed by the relative displacement from the position of the agent to the local goal $l_t$ and the current RGB observation $s^{rgb}_t$, is processed to compute an atomic action $a_t$. The available actions are: \textit{move ahead 0.25m}, \textit{turn left 10\textdegree}, \textit{turn right 10\textdegree}, with the addition of a \textit{stop} action when performing the navigation task. During training with reinforcement learning, the reward of the local policy $r^{local}_t$ encourages the agent to reduce the distance from the local goal:
\begin{equation}
r^{local}_t(\mathbf{x}_t, \mathbf{x}_{t+1}) = d(\mathbf{x}_{t}) - d(\mathbf{x}_{t+1}),
\label{eq:local_reward}
\end{equation}
where $d(\mathbf{x}_t)$ is the euclidean distance between the agent and the local goal $l_t$ at timestep $t$. Following the hierarchical design, the global goal is sampled every $N_G$ timesteps, while the local goal is reset if a new global goal is sampled, if the previous local goal is found to be in an occupied area, or if the previous local goal has been reached.

%% file: sections/05_experiments.tex
\begin{figure}[t!]
\centering
\includegraphics[width=0.98\linewidth]{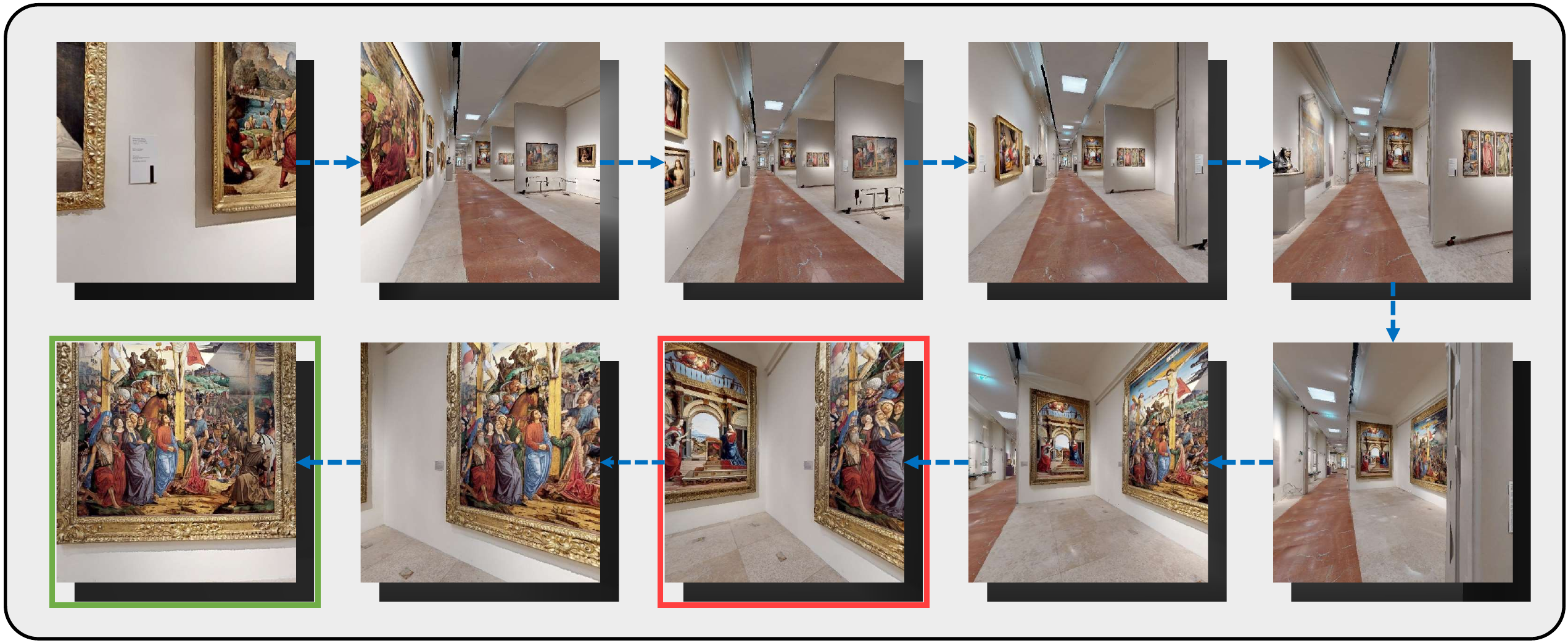}
\vspace{-0.2cm}
\caption{An episode of \POINTNAV~in \DATASETS~where consecutive frames have a distance of 10 timesteps approximately. The red frame indicates the stop action in the traditional PointNav task. The green frame corresponds to the stop action in \POINTNAV.}
\label{fig:pointnav}
\vspace{-0.3cm}
\end{figure}

We perform experiments on the proposed dataset comparing various models trained with different global rewards on another dataset, with models trained from scratch or finetuned on \DATASETS~on exploration and \POINTNAV~to evaluate the performance gap between these approaches and highlight the difference between the characteristics of AG3D compared to other datasets. A sample episode of PointGoal++ navigation of \DATASET~is shown in Fig.~\ref{fig:pointnav}.

\subsection{Experimental Setting}
\tinytit{Evaluation protocol}

The baselines are trained with Coverage, Anticipation, and Curiosity rewards on the Gibson dataset for $\approx5$M frames corresponding to 12 GPU-days on NVIDIA V100. The best performing approach among the baseline is also both trained from scratch and finetuned, but since high-quality textures and memory occupancy of \DATASETS~do not allow training with the same number of environments in parallel as Gibson, we trained the model from scratch on \DATASETS~with the same GPU time for $\approx2.8$M frames, while the finetuned model is trained for $\approx1$M additional frames.

For the exploration task we evaluate the following metrics: \textbf{IoU} (Intersection-over-Union) between the map built during at end of the episode and the ground-truth map. \textbf{Acc} measures the correctly reconstructed map in m\textsuperscript{2}. \textbf{AS} indicates the area seen by the agent during exploration (in m\textsuperscript{2}). \textbf{FIoU}, \textbf{OIoU}, \textbf{FAS}, and \textbf{OAS} measure, respectively, IoU and area seen for free and occupied portions of the environment. \textbf{TE} and \textbf{AE} are the translation and angular error between estimated and ground-truth pose measured respectively in meter and degrees.
PointGoal++ navigation is evaluated considering these metrics: \textbf{D2G} (Distance to Goal) and \textbf{OE} (Orientation Error) are the mean geodesic distance to the goal and the mean orientation error at the end of the episode. The orientation error is computed considering the vector between the center of the artwork and the position of the agent as ground-truth. \textbf{SR} (Success Rate) is the percentage of episodes terminated successfully. In \POINTNAV~the agent needs to be within 0.2 meters to the goal and with an orientation error lower than 10 degrees. \textbf{PGSR} (PointNav Success Rate) and \textbf{ASR} (Angular Success Rate) consider only one component of SR; respectively D2G and OE. \textbf{SPL} and \textbf{SoftSPL} are success rates weighted on the length of the trajectory of the agent.

\tit{Implementation details}
The experiments are performed extracting $128 \times 128$ RGB-D observations from the acquired 3D model using the Habitat simulator. The maximum length of the exploration episodes during training is set to $T=500$. Regarding the mapping process, we set $L=101$ and $G=2881$ for the local and global map dimensionalities. The action space grid $H \times H$ of the global policy is $240 \times 240$. The maximum distance of the local goal $l_t$ from the position of the agent is $D=0.5$m for exploration and $D=0.25$m for \POINTNAV.

\subsection{Experimental Results}

\input{tables/exploration}

\tinytit{Exploration results} As a first experiment, in Table~\ref{tab:exploration} we compare the considered models on the exploration task on the \DATASETS~validation set. Each exploration episode has a length of $T=1000$ timesteps during which the agent has to disclose the initially unknown environment. Among the baselines trained only on the Gibson dataset, the Coverage-based model achieves the best results in terms of IoU and Area Seen in both noise-free and noisy settings. The model trained with Coverage from scratch obtains competitive results even using fewer training frames (2.8M vs. 5M), showing the importance of adapting the models to \DATASETS. This conclusion is supported by the fact that the model trained on Gibson and finetuned for 1M frames on \DATASETS~achieves the best results on noise-free and noisy settings, with a significant margin on the second-best model. In both settings, the performance gap in terms of Area Seen (85.9m\textsuperscript{2} and 81.2m\textsuperscript{2}) and IoU (0.082 and 0.043) between the best models trained only on Gibson dataset and using \DATASETS~denotes the need of adapting the weight of the models to the different visual characteristics and occupation of \DATASETS.

\input{tables/navigation}

\tit{\POINTNAV~results} 
Moving on to the navigation task, models trained on exploration substitute the global goal with a fixed goal specified by the navigation episode.
Experimental results on \POINTNAV, shown in Table~\ref{tab:pointnav}, present a similar trend as on the exploration task. In fact, the Coverage model has the best results in terms of SPL and Success Rate related metrics among the models trained only on Gibson. Moving to the Coverage models trained on \DATASETS, in the noise-free setting, the model trained from scratch achieves the best results even in comparison to the finetuned counterpart which is trained with more than double the total observations (2.8M vs 6M). This behavior can be explained by the performance of its mapper that is trained for more frames using visual observation from \DATASETS~(2.8M vs 1M) and extracts a more detailed map sacrificing robustness and generalization. Accordingly, in the noisy setting, the higher robustness of the Coverage-based model finetuned on \DATASETS~regains the first place with a noteworthy margin on the other models, while the Coverage model trained from scratch goes down to the second position in terms of SPL and SR. As in the case of the exploration task, the performance gap between models trained on Gibson and using \DATASETS~(0.045 and 0.145 for SPL in noise-free and noisy settings) stresses the importance of adapting the parameters to the features extracted from \DATASETS. Moreover, it is worth noting that the gap of the best model from noise-free to noisy navigation (0.432 for SPL) is a consequence of the length of the navigation episodes of \DATASETS, and the difficulty of performing precise lengthy trajectories in the presence of noise. This is an interesting aspect that the \DATASETS~dataset offers for exploration in future works.

%% file: tables/exploration.tex
\begin{table}[t]
\footnotesize
\centering
\caption{Exploration results over the 100 episodes of the \DATASETS~validation split in noise-free and noisy conditions.}
\vspace{-0.1cm}
\label{tab:exploration}
\setlength{\tabcolsep}{.3em}
\resizebox{\linewidth}{!}{
\begin{tabular}{l cc ccccccccc}
\toprule
\textbf{Model} & \textbf{Training} & & \textbf{IoU} $\uparrow$ & \textbf{FIoU} $\uparrow$ & \textbf{OIoU} $\uparrow$ & \textbf{Acc} $\uparrow$ & \textbf{AS} $\uparrow$ & \textbf{FAS} $\uparrow$ & \textbf{OAS} $\uparrow$ & \textbf{TE} $\downarrow$ & \textbf{AE} $\downarrow$ \\
\midrule
\textbf{Noise-Free}\\
\hspace{0.25cm}Anticipation~\cite{ramakrishnan2020occupancy} & Gibson & & 0.163 & 0.170 & 0.157 & 294.4 & 290.6 & 258.3 & 32.3 & 0.0 & 0.0\\
\hspace{0.25cm}Curiosity~\cite{pathak2017curiosity} & Gibson & & 0.175 & 0.184 & 0.166 & 317.9 & 317.5 & 281.7 & 35.8 & 0.0 & 0.0\\
\hspace{0.25cm}Coverage~\cite{chaplot2019learning,ramakrishnan2021exploration} & Gibson & & 0.214 & 0.237 & 0.191 & 403.1 & 384.3 & 341.1 & 43.2 & 0.0 & 0.0 \\
\hspace{0.25cm}Coverage~\cite{chaplot2019learning,ramakrishnan2021exploration} & \DATASETS & & 0.219 & 0.239 & 0.200 & 400.6 & 354.6 & 316.6 & 38.0 & 0.0 & 0.0 \\
\hspace{0.25cm}Coverage~\cite{chaplot2019learning,ramakrishnan2021exploration} & Gibson+\DATASETS & & \textbf{0.296} & \textbf{0.313} & \textbf{0.278} & \textbf{531.8} & \textbf{470.2} & \textbf{418.1} & \textbf{52.1} & 0.0 & 0.0 \\
\midrule
\textbf{Noisy}\\
\hspace{0.25cm}Anticipation~\cite{ramakrishnan2020occupancy} & Gibson & & 0.144 & 0.157 & \textbf{0.131} & 269.6 & 281.7 & 249.9 & 31.9 & \textbf{0.48} & \textbf{2.95}\\
\hspace{0.25cm}Curiosity~\cite{pathak2017curiosity} & Gibson & & 0.119 & 0.151 & 0.086 & 251.1 & 307.3 & 272.8 & 34.5 & 2.62 & 15.99\\
\hspace{0.25cm}Coverage~\cite{chaplot2019learning,ramakrishnan2021exploration} & Gibson & & 0.148 & 0.203 & 0.093 & 327.0 & 380.7 & 337.9 & 42.8 & 2.98 & 12.54\\
\hspace{0.25cm}Coverage~\cite{chaplot2019learning,ramakrishnan2021exploration} & \DATASETS & & 0.144 & 0.200 & 0.088 & 320.0 & 356.4 & 317.3 & 39.2 & 2.65 & 13.35 \\
\hspace{0.25cm}Coverage~\cite{chaplot2019learning,ramakrishnan2021exploration} & Gibson+\DATASETS & & \textbf{0.191} & \textbf{0.266} & 0.116 & \textbf{427.3} & \textbf{461.9} & \textbf{413.2} & \textbf{48.7} & 2.68 & 10.16 \\
\bottomrule
\end{tabular}
}
\vspace{-3mm}
\end{table}

%% file: tables/navigation.tex
\begin{table}[t]
\footnotesize
\centering
\caption{\POINTNAV~results on the \DATASETS~navigation episodes under noise-free and noisy settings.}
\vspace{-0.1cm}
\label{tab:pointnav}
\setlength{\tabcolsep}{.3em}
\resizebox{\linewidth}{!}{
\begin{tabular}{l cc cccccccc}
\toprule
\textbf{Model} & \textbf{Training} & & \textbf{SPL} $\uparrow$ & \textbf{SoftSPL} $\uparrow$ & \textbf{SR} $\uparrow$ & \textbf{PNSR} $\uparrow$ & \textbf{ASR} $\uparrow$ & \textbf{Steps} $\downarrow$ & \textbf{D2G} $\downarrow$ & \textbf{OE} $\downarrow$ \\
\midrule
\textbf{Noise-Free}\\
\hspace{0.25cm}Anticipation~\cite{ramakrishnan2020occupancy} & Gibson & & 0.697 & 0.780 & 0.803 & 0.873 & 0.808 & 364.3 & 4.131 & 12.2 \\
\hspace{0.25cm}Curiosity~\cite{pathak2017curiosity} & Gibson & & 0.625 & 0.706 & 0.732 & 0.803 & 0.732 & 416.4 & 7.934 & 17.0\\
\hspace{0.25cm}Coverage~\cite{chaplot2019learning,ramakrishnan2021exploration} & Gibson & & 0.760 & 0.838 & 0.876 & 0.954 & 0.883 & 314.6 & 0.700 & 5.2 \\
\hspace{0.25cm}Coverage~\cite{chaplot2019learning,ramakrishnan2021exploration} & \DATASETS & & \textbf{0.805} & \textbf{0.875} & \textbf{0.898} & \textbf{0.973} & \textbf{0.908} & \textbf{270.3} & \textbf{0.268} & \textbf{4.8} \\
\hspace{0.25cm}Coverage~\cite{chaplot2019learning,ramakrishnan2021exploration} & Gibson+\DATASETS & & 0.793 & 0.873 & 0.883 & 0.964 & 0.891 & 273.1 & 0.323 & 5.0 \\
\midrule
\textbf{Noisy}\\
\hspace{0.25cm}Anticipation~\cite{ramakrishnan2020occupancy} & Gibson & & 0.211 & 0.788 & 0.224 & 0.255 & 0.387 & 338.6 & 3.152 & 32.2 \\
\hspace{0.25cm}Curiosity~\cite{pathak2017curiosity} & Gibson & & 0.225 & 0.655 & 0.243 & 0.275 & 0.341 & 446.3 & 9.746 & 38.7 \\
\hspace{0.25cm}Coverage~\cite{chaplot2019learning,ramakrishnan2021exploration} & Gibson & & 0.228 & 0.783 & 0.243 & 0.260 & 0.392 & 348.6 & 3.165 & 34.1 \\
\hspace{0.25cm}Coverage~\cite{chaplot2019learning,ramakrishnan2021exploration} & \DATASETS & & 0.235 & 0.832 & 0.248 & 0.273 & 0.445 & 306.2 & 2.420 & 28.8 \\
\hspace{0.25cm}Coverage~\cite{chaplot2019learning,ramakrishnan2021exploration} & Gibson+\DATASETS & & \textbf{0.373} & \textbf{0.853} & \textbf{0.399} & \textbf{0.443} & \textbf{0.543} & \textbf{283.8} & \textbf{1.430} & \textbf{19.8} \\
\bottomrule
\end{tabular}
}
\vspace{-3mm}
\end{table}

%% file: sections/06_conclusion.tex
In this work, we introduced the \DATASETS~photorealistic 3D dataset for embodied exploration and PointGoal navigation tasks. The dataset has been collected in an art gallery, which features larger and more uncluttered spaces compared to most of the environments available in commonly used benchmark datasets. For the PointNav task, we propose a variant that is more suitable to the type of environment in the \DATASETS~dataset. The variant entails not only reaching the specified coordinates, as in standard PointNav but also assuming a specified orientation. We also present an experimental comparison of state-of-the-art approaches on the devised dataset, which can serve as baselines for future research on embodied AI tasks performed in museum-like environments.